\documentclass[conference]{IEEEtran}
\IEEEoverridecommandlockouts
\usepackage{cite}
\usepackage{amsmath,amssymb,amsfonts}
\usepackage{algorithmic}
\usepackage{graphicx}
\usepackage[caption=false]{subfig}
\usepackage{textcomp}
\usepackage{multirow}
\usepackage{booktabs}
\def\BibTeX{{\rm B\kern-.05em{\sc i\kern-.025em b}\kern-.08em
    T\kern-.1667em\lower.7ex\hbox{E}\kern-.125emX}}

\makeatother

\begin{document}

\title{Mitigating Gender Bias in Depression Detection via Counterfactual Inference
\thanks{* Corresponding author.}
}

\author{\IEEEauthorblockN{Mingxuan Hu} 
\IEEEauthorblockA{\textit{School of Advanced Technology} \\
\textit{Xi’an Jiaotong-Liverpool University}\\
Suzhou, China \\
Mingxuan.Hu22@student.xjtlu.edu.cn}
\\
\IEEEauthorblockN{Ziqi Liu} 
\IEEEauthorblockA{\textit{School of Advanced Technology} \\
\textit{Xi’an Jiaotong-Liverpool University}\\
Suzhou, China \\
Ziqi.Liu22@student.xjtlu.edu.cn}
\and
\IEEEauthorblockN{Hongbo Ma} 
\IEEEauthorblockA{\textit{School of Advanced Technology} \\
\textit{Xi’an Jiaotong-Liverpool University}\\
Suzhou, China \\
Hongbo.Ma22@student.xjtlu.edu.cn}
\\
\IEEEauthorblockN{Jiaqi Liu} 
\IEEEauthorblockA{\textit{School of Advanced Technology} \\
\textit{Xi’an Jiaotong-Liverpool University}\\
Suzhou, China \\
Jiaqi.Liu03@xjtlu.edu.cn}
\and
\IEEEauthorblockN{Xinlan Wu} 
\IEEEauthorblockA{\textit{School of Advanced Technology} \\
\textit{Xi’an Jiaotong-Liverpool University}\\
Suzhou, China \\
Xinlan.Wu23@student.xjtlu.edu.cn}
\\
\IEEEauthorblockN{Yangbin Chen*} 
\IEEEauthorblockA{\textit{School of Advanced Technology} \\
\textit{Xi’an Jiaotong-Liverpool University}\\
Suzhou, China \\
Yangbin.Chen@xjtlu.edu.cn}
}

\maketitle

\begin{abstract}
Audio-based depression detection models have demonstrated promising performance but often suffer from gender bias due to imbalanced training data. Epidemiological statistics show a higher prevalence of depression in females, leading models to learn spurious correlations between gender and depression. Consequently, models tend to over-diagnose female patients while underperforming on male patients, raising significant fairness concerns. To address this, we propose a novel Counterfactual Debiasing Framework grounded in causal inference. We construct a causal graph to model the decision-making process and identify gender bias as the direct causal effect of gender on the prediction. During inference, we employ counterfactual inference to estimate and subtract this direct effect, ensuring the model relies primarily on authentic acoustic pathological features.  Extensive experiments on the DAIC-WOZ dataset using two advanced acoustic backbones demonstrate that our framework not only significantly reduces gender bias but also improves overall detection performance compared to existing debiasing strategies.
\end{abstract}


\section{Introduction}
Depression is a serious mental disorder that impairs cognitive and behavioral functions and increases suicide risk \cite{goldman1999awareness}. The World Health Organization \cite{who2025report} reported a global prevalence of about 4\% in 2021, causing an estimated annual productivity loss of one trillion dollars. Therefore, early depression diagnosis is critically important. Traditional diagnostic methods mainly rely on subjective scales such as the Patient Health Questionnaire-8 (PHQ-8) \cite{kroenke2009phq} and the Beck Depression Inventory (BDI) \cite{beck1961inventory}, as well as structured interviews by clinicians \cite{endicott1978diagnostic}. However, due to limited medical expertise and healthcare resources \cite{goldman1999awareness}, many patients struggle to receive timely diagnosis. 

In recent years, automated depression detection based on artificial intelligence has gained increasing attention. Among various modalities, audio-based methods show great potential due to easy data acquisition, low cost, and non-invasiveness. Models like wav2vec 2.0 \cite{baevski2020wav2vec} have been successfully applied to learn effective speech representations, forming a strong foundation for high-performance depression detection.

However, deep learning model performance heavily depends on training data quality and distribution. A crucial but often overlooked issue is that these models can unintentionally learn and amplify inherent data biases. In depression research, epidemiological data show female depression prevalence is about twice that of males \cite{kuehner2017depression}, causing a gender imbalance in datasets, where female samples, especially those diagnosed with depression, significantly exceeding the number of male samples. This bias may lead models to rely more on gender information rather than true pathological voice features in depression detection. Consequently, the model tends to over-diagnose female patients \cite{floyd1997problems} while exhibiting lower recognition accuracy for male patients \cite{shi2021hypothesis}, which raises significant fairness and ethical concerns. Given the serious consequences, it is crucial to adopt effective mitigation methods.

Regarding the issue of gender bias, researchers have proposed several mitigation strategies, most of which are pre-processing (data-level) approaches. For example, Bailey and Plumbley \cite{bailey2021gender} enforced a balanced number of samples across different classes in the training set by sub-sampling the majority group. In addition, Cheong et al. \cite{cheong2023s} proposed a feature-level data augmentation technique to synthesize new virtual samples. However, these methods only adjust superficial data distribution and fail to break the spurious causal link between gender and depression within the model’s decision process. Therefore, more systematic approachs are needed to eliminate such bias.

Inspired by causal inference theory \cite{pearl2010introduction} and counterfactual inference theory \cite{pearl2011logic}, and drawing on their successful application in eliminating language bias in the Visual Question Answering (VQA) domain \cite{niu2021counterfactual}, we propose a novel \textbf{Counterfactual Debiasing Framework}. We formally define gender bias as the direct causal effect of gender information on depression status. During model training, we introduce counterfactual interventions aimed at quantifying this bias effect and removing it from the model’s overall judgment during final inference. This results in a more equitable diagnosis that relies more heavily on authentic acoustic pathological features. Overall, the main contributions of this paper are as follows:

\begin{enumerate}
    \item Analyze and quantify the gender bias present in automatic depression detection.
    \item Propose a novel end-to-end counterfactual debiasing framework that cuts off the spurious causal link between gender and depression.
    \item Conduct experiments on two advanced acoustic baseline models, with results demonstrating that the proposed framework not only enhances overall performance, but also improves model fairness.
\end{enumerate}

\section{Related Work}
\subsection{Audio-based Depression Detection}
Early audio-based depression detection primarily relied on handcrafted acoustic features and traditional machine learning classifiers \cite{li2025automated}, such as landmark words with Support Vector Machine (SVM) \cite{huang2019natural}, and eGeMAPS features with logistic regression \cite{jayawardena2020ordinal}. With the rise of deep learning, AI chatbots have been employed to collect audio signals \cite{xu2025deep}, and end-to-end deep speech features often outperform traditional acoustic features \cite{chen2025speech}. Niu et al. \cite{niu2020multimodal} proposed the Spatio-Temporal Attention (STA) network, which captures both spatial and temporal features of speech signals, and highlights key frames through an attention mechanism. To address the issue of variable-length speech segments in clinical interviews, Shen et al. \cite{shen2022automatic} introduced a framework based on the NetVLAD aggregation module, which encodes acoustic feature sequences of arbitrary length into fixed-dimensional vector representations. Building on the works of \cite{niu2020multimodal} and \cite{shen2022automatic}, we construct two advanced acoustic backbone models to comprehensively evaluate the proposed debiasing framework.

\subsection{Debiasing Methods in Depression Detection}
Existing debiasing methods for depression detection fall into three types: pre-processing, in-processing, and post-processing \cite{cheong2021hitchhiker}. Pre-processing methods operate at the data level, such as sub-sampling the majority group \cite{bailey2021gender}, or applying data augmentation for the minority group \cite{cheong2023s}. However, sub-sampling may reduce information diversity, while data augmentation risks introducing unrealistic or low-quality data. In-processing methods intervene during model training, with adversarial training \cite{zhang2024novel} as a typical example. Though promising, it often suffers from instability and convergence issues, and its “black-box” mechanism makes it hard to interpret what bias is actually removed. Post-processing methods adjust predictions after inference, like setting different classification thresholds per gender \cite{dang2024fairness}, but only corrects final outputs without addressing biased feature representations inside the model.

The common limitation of these traditional debiasing methods is that they address bias only at the data level or model output level, neglecting the model’s internal decision process and underlying causes. In contrast, our counterfactual debiasing framework, grounded in causal inference, accurately identifies and removes spurious gender-related causal paths, enabling more effective and interpretable bias mitigation.

\subsection{Causal Inference and Counterfactual Inference}
Causal inference \cite{pearl2010introduction} is a theoretical framework for analyzing causal relationships between variables. It goes beyond simple correlation analysis by formally identifying and removing confounding effects using models such as causal graphs. Counterfactual inference \cite{pearl2011logic}, as a key branch of causal inference, constructs hypothetical virtual scenarios under interventions to quantify bias effects.

Recently, causal inference has been applied in various cutting-edge AI fields, such as addressing social bias in large language models \cite{liu2025large}, and tackling popularity bias in recommender systems \cite{luo2024survey}. Furthermore, a study on VQA forms the theoretical foundation of our research. Niu et al. \cite{niu2021counterfactual} found that due to imbalanced training samples, VQA models often ignore image content and rely solely on language priors, causing language bias. To separate this bias, they used counterfactual inference to estimate the model’s predictions when the image is unseen during training, and removed the bias during inference, encouraging the model to depend more on true visual information. Following this idea, we innovatively introduce counterfactual inference into audio-based depression detection to separate and eliminate gender bias.

\section{Methodology}
\subsection{Basic Concepts of Causal Inference}
In this section, we introduce some basic concepts of causal inference \cite{pearl2010introduction},\cite{pearl2011logic},\cite{niu2021counterfactual}. We use capital letters (e.g., $X$) to represent variables and corresponding lowercase letters (e.g., $x$ and $\bar{x}$) to represent the state of the variables.

\begin{figure}[!t]
    \centering
    \subfloat[]{%
        \includegraphics[width=0.32\linewidth]{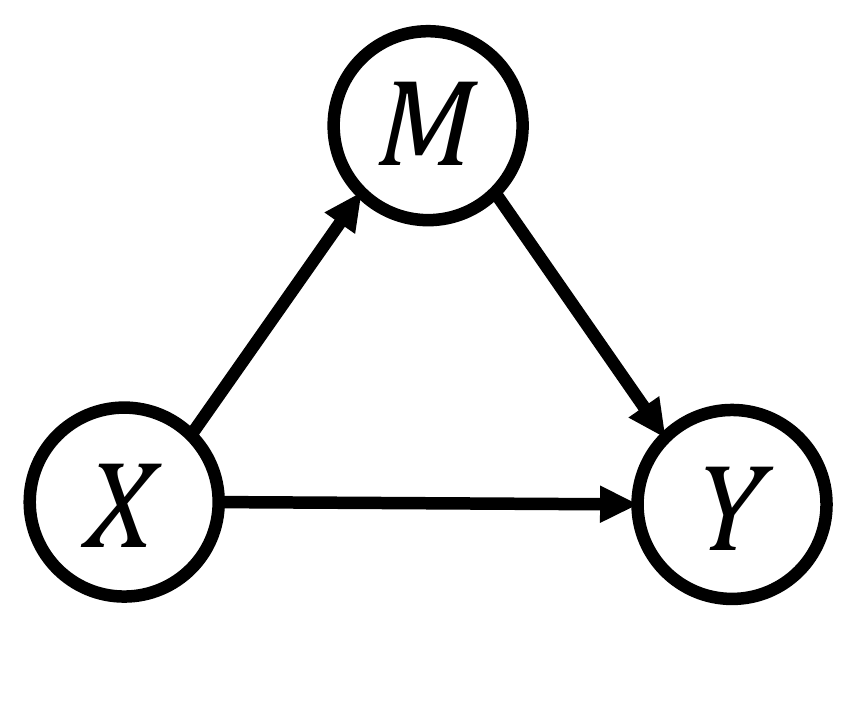}%
        \label{fig3-sub1}%
    }
    \hspace{1pt}
    \subfloat[]{%
        \includegraphics[width=0.32\linewidth]{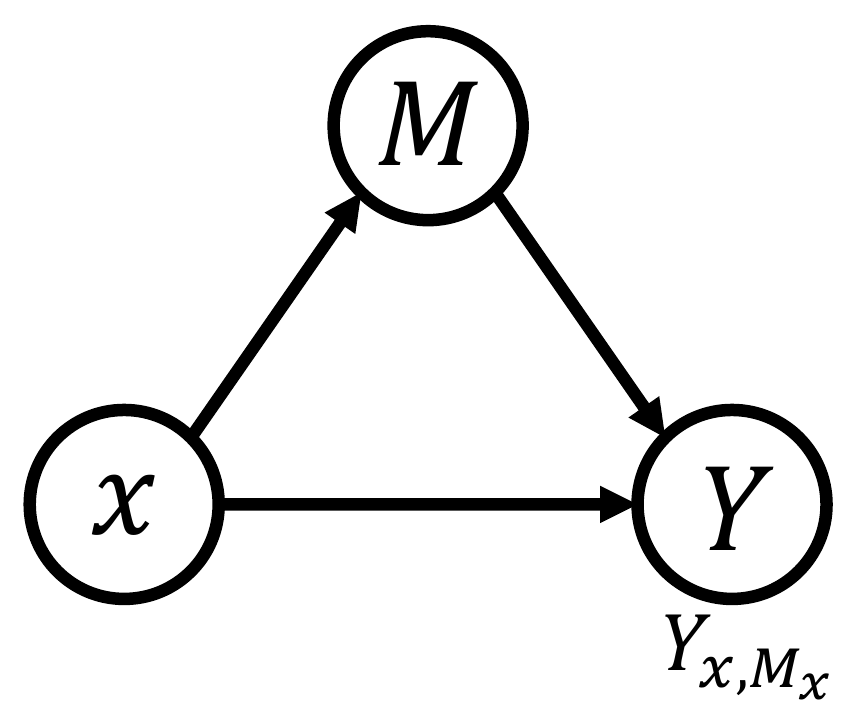}%
        \label{fig3-sub2}%
    }
    \hspace{1pt}
    \subfloat[]{%
        \includegraphics[width=0.32\linewidth]{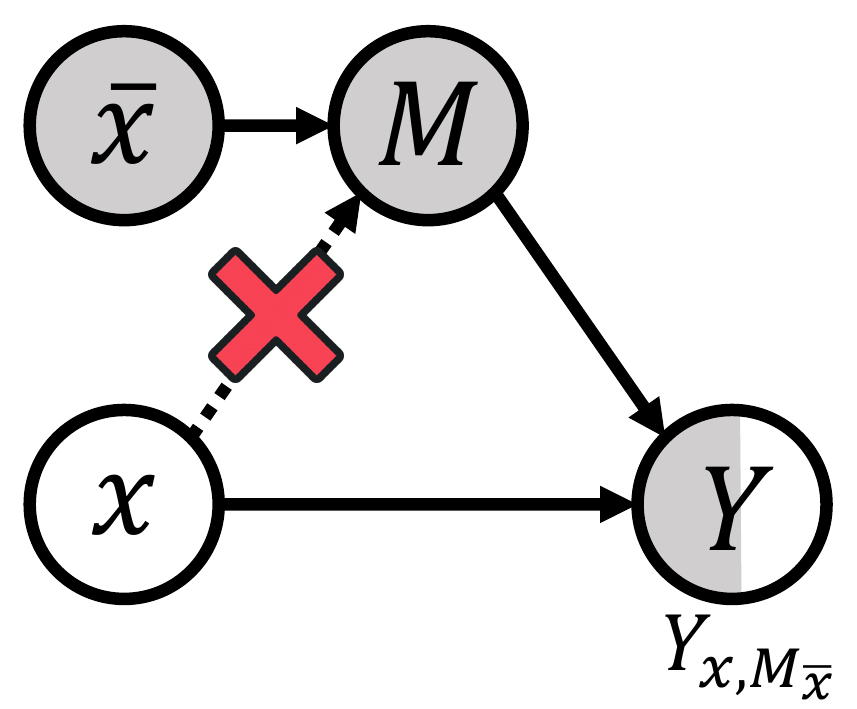}%
        \label{fig3-sub3}%
    }
    
    \caption{(a) Example of causal graph. (b) Example of factual scenario. (c) Example of counterfactual scenario. White nodes are at the state under $X = x$, while gray nodes are at the state under $X = \bar{x}$.}
    \label{fig3}
\end{figure}

\textbf{Causal graph} is a directed acyclic graph used to model causal relationships between variables, denoted as $\mathcal{G} = \{\mathcal{V}, \mathcal{E}\}$, where $\mathcal{V}$ denotes variables and $\mathcal{E}$ denotes causal relationships. Fig.~\ref{fig3}\subref{fig3-sub1} is an example of causal graph, where the total causal effect of $X$ on $Y$ consists of a direct effect (i.e., $X \to Y$) and an indirect effect (i.e., $X \to M \to Y$).

\textbf{Factual scenario} represents the observed situation in reality. Fig.~\ref{fig3}\subref{fig3-sub2} shows an example of factual scenario, where $X$ is in state $x$. Given $M$ is causally affected by $X$, the state of $M$ under $X = x$ can be represented as $M_x = M(X = x)$. Since $Y$ is causally affected by both $X$ and $M$ (i.e., directly and indirectly by $X$), the state of $Y$ under $X = x$ can be represented as:

\begin{equation}
    Y_{x, M_{x}} = Y(X = x, M = M(X = x))
\end{equation}

\textbf{Counterfactual scenario} is a virtual situation arising from intervening on certain variable states in the factual scenario. Fig.~\ref{fig3}\subref{fig3-sub3} is the counterfactual scenario corresponding to the factual scenario in Fig.~\ref{fig3}\subref{fig3-sub2}. Specifically, $X$ is in state $x$, while $M$ is intervened to take the state it would have under $X = \bar{x}$ (Note that $X$ can be simultaneously set to different states $x$ and $\bar{x}$ only in the counterfactual scenario). In this case, the state of $Y$ can be represented as:

\begin{equation}
    Y_{x, M_{\bar{x}}} = Y(X = x, M = M(X = \bar{x}))
\end{equation}

\textbf{Total Indirect Effect (TIE)} represents the causal effect of $X$ on $Y$ through the mediator $M$. It is defined by comparing the factual scenario shown in Fig.~\ref{fig3}\subref{fig3-sub2} and the counterfactual scenario shown in Fig.~\ref{fig3}\subref{fig3-sub3}. Specifically, TIE is the difference in the state of $Y$ when $M$ changes from the state under $X = x$ to the state under $X = \bar{x}$, while $X$ is fixed in state $x$. It can be expressed as:

\begin{equation}
    TIE = Y_{x, M_{x}} - Y_{x, M_{\bar{x}}}
\end{equation}

The intuition behind TIE is to isolate the effect transmitted solely through the mediator $M$. In other words, if we treat the direct effect $X \to Y$ as bias, then TIE measures the unbiased effect in the system.

\subsection{Causal Modeling on Depression Detection}
Following common practice, we define audio-based depression detection as a binary classification task, where the model predicts the depression state $d$ from the candidate set $D = \{\text{Depressed}, \text{Non-depressed}\}$ using acoustic cues $C = c$ and gender information $G = g$.

\begin{figure}[!t]
    \centering
    \subfloat[]{
        \includegraphics[width=0.29\linewidth]{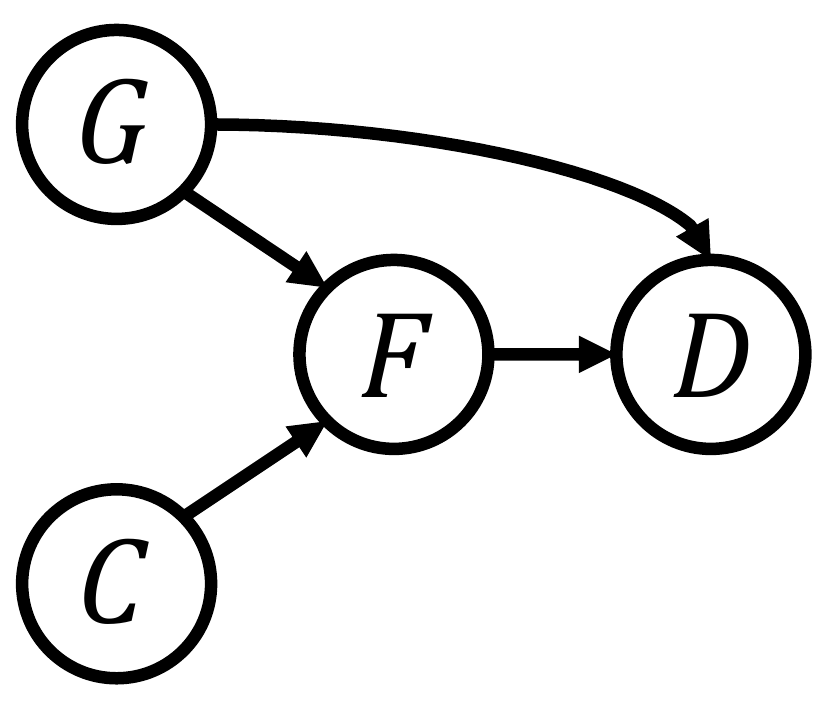}
        \label{fig4-sub1}
    }
    \hfill
    \subfloat[]{
        \includegraphics[width=0.65\linewidth]{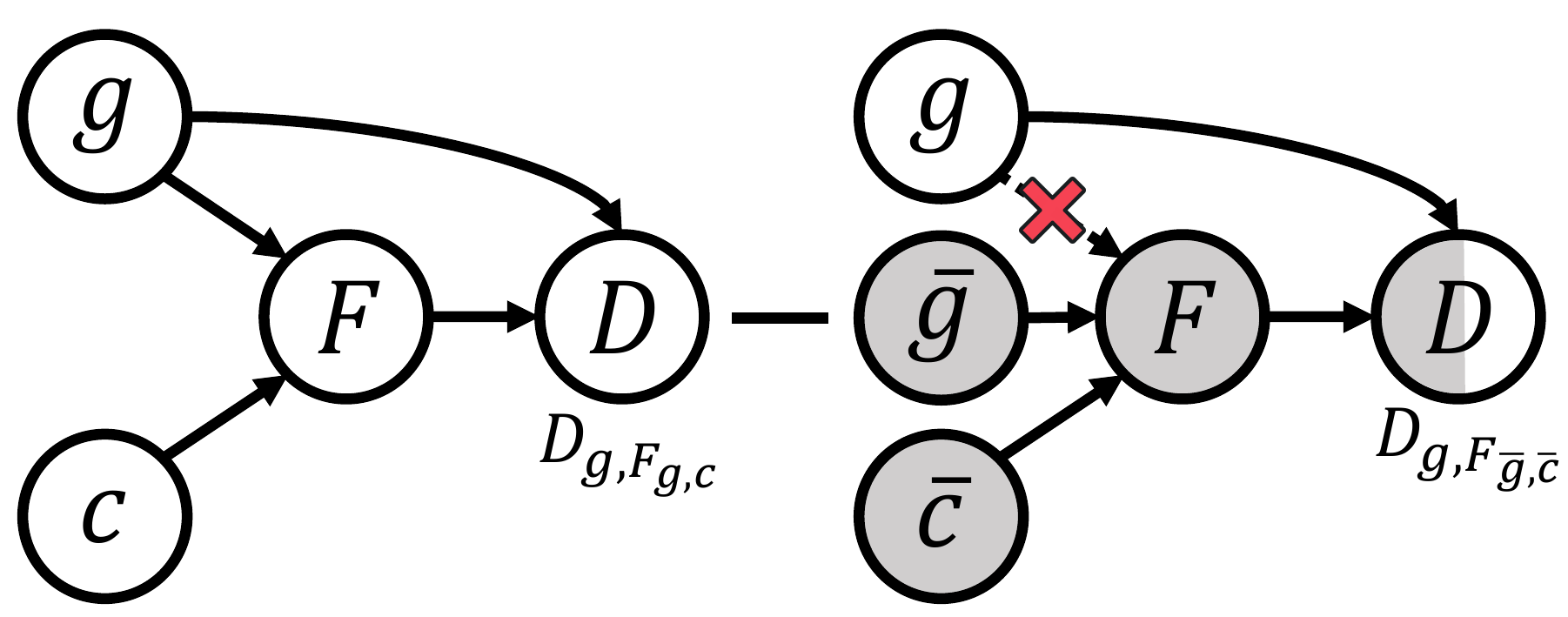}
        \label{fig4-sub2}
    }
    
    \caption{(a) Causal graph for depression detection. (b) Comparison between factual and counterfactual scenarios in depression detection. White nodes are at the state under $G = g$ and $C = c$, while gray nodes are at the state under $G = \bar{g}$ and $C = \bar{c}$.}
    \label{fig4}
\end{figure}

The causal graph of our task is shown in Fig.~\ref{fig4}\subref{fig3-sub1}. The total effect of $G$ and $C$ on $D$ can be decomposed into direct and indirect effects. We aim to fuse $G$ and $C$ into an unbiased feature F that truly captures the pathological information to indirectly affect D (i.e., $G, C \to F \to D$). However, as previously discussed, there is a spurious causal link from $G$ to $D$, causing $G$ to directly affect $D$ (i.e., $G \to D$), which represents the gender bias we want to mitigate.

To achieve this goal, we compare the related factual and counterfactual scenarios, illustrated in Fig.~\ref{fig4}\subref{fig4-sub2}. In factual scenario, $G$ is in state $g$ and $C$ in state $c$, the state of $D$ can be represented as:

\begin{equation}
    D_{g, F_{g, c}} = D(G = g, F = F(G = g, C = c))
\end{equation}

By intervening, we obtain the corresponding counterfactual scenario, with $G$ in state $g$ while $F$ is intervened to take the state it would have under $G = \bar{g}$ and $C = \bar{c}$. Since the response of mediator $F$ to inputs is blocked, model relies solely on the gender information for prediction, and the state of $D$ can be represented as:

\begin{equation}
    D_{g, F_{\bar{g}, \bar{c}}} = D(G = g, F = F(G = \bar{g}, C = \bar{c}))
\end{equation}

Thus, the mitigation of gender bias can be achieved by subtracting the direct effect $G \to D$ from the total effect, resulting in the TIE. We use TIE for final debiased inference:

\begin{equation}
    TIE = D_{g, F_{g, c}} - D_{g, F_{\bar{g}, \bar{c}}}
\end{equation}

\subsection{Counterfactual Debiasing Framework}
\textbf{Parameterization}. The structure of our proposed counterfactual debiasing framework is shown in Fig.~\ref{fig3}. The computation of $D_{g, F_{g, c}}$ is parameterized by two neural models $\mathcal{M}_G$ and $\mathcal{M}_F$, along with a fusion function $h$:

\begin{equation}
\begin{aligned}
    D_g = \mathcal{M}_G(g), & \quad D_{F_{g, c}} = \mathcal{M}_F(g,c),\\D_{g, F_{g, c}} &= h(D_g, D_{F_{g, c}})
\end{aligned}
\end{equation}

where $\mathcal{M}_G$ is a gender-only model simulating $G \to D$, and $\mathcal{M}_F$ is a gender-acoustic fusion model simulating $G, C \to F \to D$. We adopt the additive nonlinear fusion method from \cite{niu2021counterfactual} to combine $D_g$ and $D_{F_{g, c}}$ into $D_{g, F_{g, c}}$:

\begin{equation}
    h(D_g, D_{F_{g, c}}) = log \,\sigma(D_g + D_{F_{g, c}})
\end{equation}

As previously discussed, in the counterfactual scenario, $F$ takes the state under $G = \bar{g}$ and $C = \bar{c}$ (i.e., no gender or acoustic cues). Since neural networks cannot handle empty inputs, we assume the model randomly guesses the most reasonable output when given empty input, represented by a global learnable parameter $\varepsilon$. Thus, under no gender or acoustic cues, logit $D_{F_{\bar{g}, \bar{c}}}$ is expressed as:

\begin{equation}
    D_{F_{\bar{g}, \bar{c}}} = \mathcal{M}_F(\bar{g}, \bar{c}) = \mathcal{M}_F(\varepsilon)
\end{equation}

\textbf{Training.} The framework jointly optimizes $\mathcal{M}_G$, $\mathcal{M}_F$, and $\varepsilon$ by minimizing a total loss composed of classification loss $\mathcal{L}_{cls}$ and Kullback-Leibler (KL) divergence loss $\mathcal{L}_{kl}$.

$\mathcal{L}_{cls}$ ensures that $\mathcal{M}_G$ learns sufficient gender bias, while $\mathcal{M}_F$ learns effective information without bias. It is defined as:

\begin{equation}
    \mathcal{L}_{cls} = \mathcal{L}_{M_{G}}(D_g,d) + \mathcal{L}_{M_{F}}(D_{g, F_{g, c}},d)
\end{equation}

where $\mathcal{L}_{M_{G}}$ and $\mathcal{L}_{M_{F}}$ are the standard cross-entropy loss functions for the logits output by $\mathcal{M}_G$ and $\mathcal{M}_F$, respectively, and the gradients from $\mathcal{L}_{cls}$ are used to update the weights of $\mathcal{M}_G$ and $\mathcal{M}_F$.

$\mathcal{L}_{kl}$ ensures the model can achieve automated debiasing. To make the model’s randomly guessed result distribution under counterfactual scenario as close as possible to the true distribution, we use $\mathcal{L}_{kl}$ to enforce this similarity:

\begin{equation}
    \mathcal{L}_{kl} = \frac{1}{|D|}\sum_{d \in D} -p(d|g,c)\,log \,p(d|g,\bar{c})
\end{equation}

where $p(d|g,c) = \text{softmax}(D_{g, F_{g, c}})$ and $p(d|g,\bar{c}) = \text{softmax}(D_{g,F_{\bar{g}, \bar{c}}})$. When minimizing $\mathcal{L}_{kl}$, only the learnable parameters $\varepsilon$ are updated, enabling automatic parameter tuning. Overall, the total training loss $\mathcal{L}$ is: 

\begin{equation}
    \mathcal{L} = \mathcal{L}_{cls} + \mathcal{L}_{kl}
\end{equation}

\textbf{Inference.} We use the debiased causal effect TIE for inference, which is implemented as:

\begin{equation}
\begin{split}
TIE &= D_{g, F_{g, c}} - D_{g, F_{\bar{g}, \bar{c}}} \\
&= h(\mathcal{M}_G(g), \mathcal{M}_F(g, c)) - h(\mathcal{M}_G(g), \mathcal{M}_F(\epsilon))
\end{split}
\end{equation}

\begin{figure}[!tbp]
    \centering
    \includegraphics[width=\columnwidth]{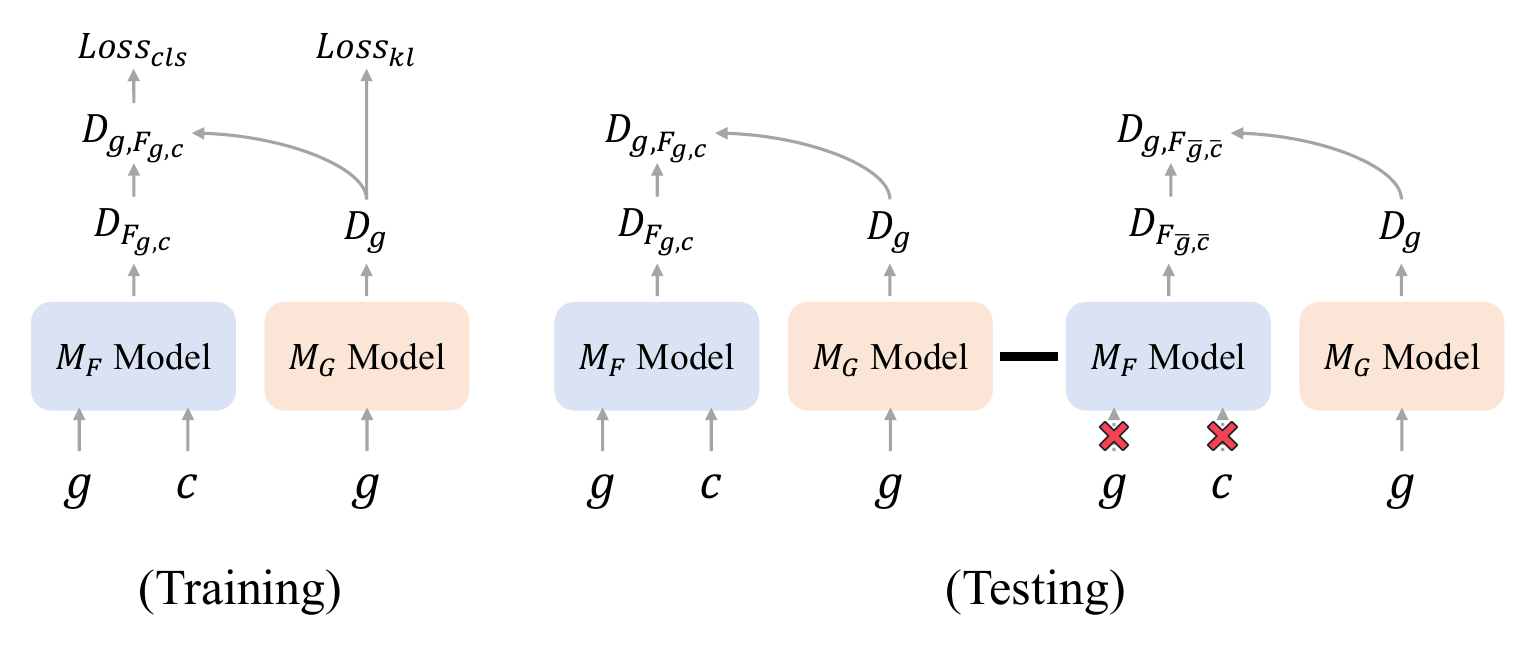}
    \caption{Architecture of Proposed Counterfactual Debiasing Framework. The framework consists of a training stage that jointly optimizes a gender-only model $\mathcal{M}_G$ and a fusion model $\mathcal{M}_F$, and a testing stage where the bias effect is subtracted from the total effect via a counterfactual path.}
    \label{fig3}
\end{figure}

\section{Experimental Setup}
\subsection{Dataset}
We used the Distress Analysis Interview Corpus - Wizard Of Oz (DAIC-WOZ) dataset \cite{gratch2014distress}, which was collected through clinical interviews conducted by a virtual agent named Ellie, to support diagnosis of mental health conditions such as depression and anxiety. For our experiments, we utilized the audio recordings, corresponding text transcriptions, and participants’ PHQ-8 questionnaire results.

Since our counterfactual debiasing framework requires no hyperparameter tuning, we combined the original training and development sets to form a new training set. Table~\ref{tab1} shows the label distribution of the combined training set. Among 63 female samples, 38.1\% (24 out of 63) were diagnosed with depression, compared to 24.1\% (19 out of 79) of male samples. This indicates that the depression rate in females is approximately 1.58 times that of males, reflecting a certain degree of gender imbalance in real-world data.

\begin{table}[!tbp]
    \centering
    \caption{Distribution of Combined DAIC-WOZ Training Set}
    \begin{tabular}{cccc}
    \hline
        ~ & Non-depressed & Depressed & Total \\ \hline
        Female & 39 (27.5\%) & 24 (16.9\%) & 63 (44.4\%) \\ 
        Male & 60 (42.3\%) & 19 (13.4\%) & 79 (55.6\%) \\ 
        Total & 99 (69.7\%) & 43 (30.3\%) & 142 (100.0\%) \\ \hline
    \end{tabular}
    \label{tab1}
\end{table}

The label distribution of the testing set is shown in Table~\ref{tab2}. Unlike the training set, the depression rates for females and males are balanced at 29.2\% (7 out of 24) and 30.4\% (7 out of 25), respectively. Therefore, if the model relies solely on gender bias learned from the training data, its performance will significantly drop on this balanced test set. Thus, the test set effectively evaluates bias and validates the effectiveness of our debiasing method.

\begin{table}[!tbp]
    \centering
    \caption{Distribution of DAIC-WOZ Testing Set}
    \begin{tabular}{cccc}
    \hline
        ~ & Non-depressed & Depressed & Total \\ \hline
        Female & 17 (36.2\%) & 7 (14.9\%) & 24 (51.1\%) \\ 
        Male & 16 (34.0\%) & 7 (14.9\%) & 23 (48.9\%) \\ 
        Total & 33 (70.2\%) & 14 (29.8\%) & 47 (100.0\%) \\ \hline
    \end{tabular}
    \label{tab2}
\end{table}

\subsection{Evaluation Metrics}
We evaluate overall model performance using F1-score, Accuracy, and Recall. To analyze gender-specific performance, we also report Male-F1 and Female-F1.

Following \cite{cheong2023s}, we introduce two fairness metrics. \textbf{Equal Accuracy (EA)} measures the absolute accuracy gap between the majority group (female, denoted as $G = 1$) and the minority group (male, denoted as $G = 0$). Values closer to 0 indicate greater fairness.

\begin{equation}
    EA = \left | Acc(G = 1) - Acc(G = 0) \right | 
\end{equation}

Additionally, \textbf{Disparate Impact (DI)} measures the ratio of positive outcome (depression, denoted as $\hat{Y} = 1$) for both the majority and minority group. Values closer to 1 indicate greater fairness.

\begin{equation}
    DI = \frac{Pr(\hat{Y} = 1 | G = 1)}{Pr(\hat{Y} = 1 | G = 0)} 
\end{equation}

\subsection{Backbone Architectures}
\begin{figure}[!tbp]
    \centering
    \includegraphics[width=\columnwidth]{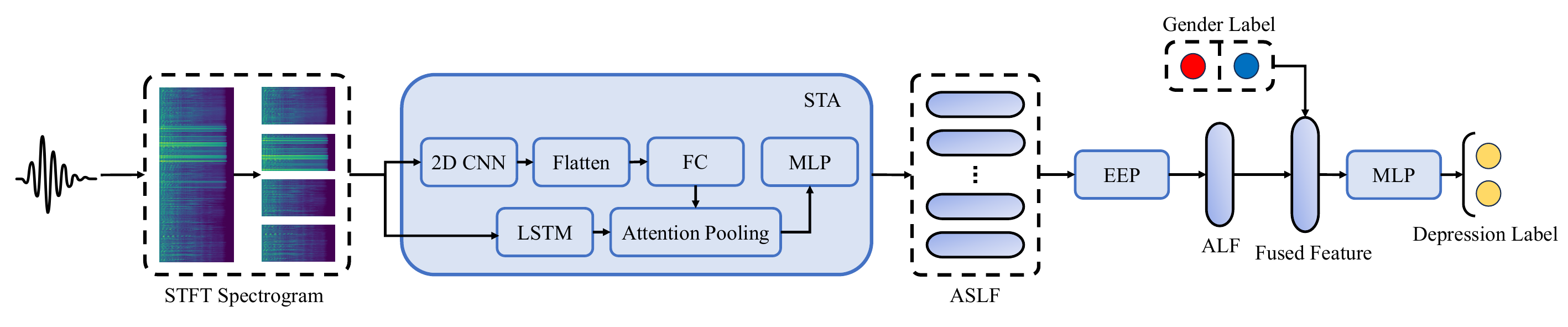}
    \caption{Architecture of STA-based backbone model. STA network has two branches, including a 2D CNN for spatial features and an LSTM for temporal dynamics. Their outputs fuse via Attention Pooling to highlight key speech frames. EEP uses eigendecomposition to capture temporal correlations and identify principal components for weighted sequence aggregation.}
    \label{fig1}
\end{figure}

We designed two different backbone models for $\mathcal{M}_F$. The first STA-based backbone model is based on \cite{niu2020multimodal}, 
illustrated in Fig.~\ref{fig1}. In preprocessing, raw audio was resampled to 8kHz and converted into 129-dimensional normalized spectrograms via Short-Time Fourier Transform (STFT). These spectrograms were segmented into fixed-length clips of 64 frames with a 32-frame stride, as STA network inputs. The network extracted Audio Segment-Level Features (ASLF) from each clip, which were aggregated into a single 64-dimensional Audio-Level Feature (ALF) using Eigen Evolution Pooling (EEP). The ALF was then concatenated with 1-dimensional gender information to form a 65-dimensional fused feature, fed into a Multi-Layer Perceptron (MLP) for binary depression classification.

\begin{figure}[!tbp]
    \centering
    \includegraphics[width=\columnwidth]{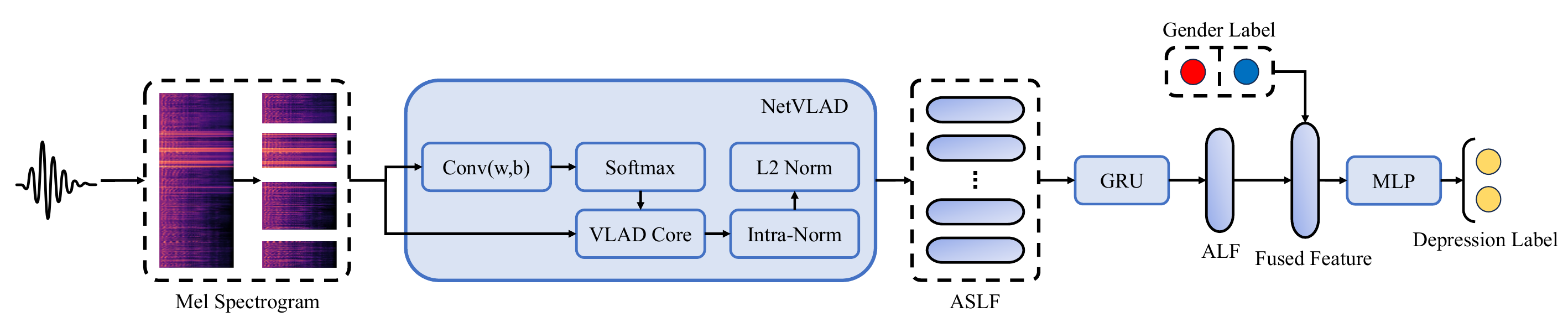}
    \caption{Architecture of the NetVLAD-based backbone model. NetVLAD module assigns local acoustic features to learnable cluster centers (VLAD Core) and sums weighted residuals to produce fixed-dimensional vectors. GRU network uses internal gating mechanism to selectively update its hidden state at each step, with the final state summarizing the sequence.}
    \label{fig2}
\end{figure}

Fig.~\ref{fig2} shows the second backbone model based on \cite{shen2022automatic}. Compared to the first, raw audio was segmented by transcript into question-level clips, producing Mel-spectrograms of varying sizes. These Mel-spectrograms passed through a NetVLAD module to extract fixed-length 256-dimensional ASLF features. ASLFs are then input to a Gated Recurrent Unit (GRU) network, whose final hidden state served as a 256-dimensional ALF feature. Subsequent steps follow the first backbone model. Since both backbone models for $\mathcal{M}_F$ use a downstream MLP for final depression detection, $\mathcal{M}_G$ is also designed as a MLP.

\subsection{Baselines}
We selected two classic debiasing methods as baselines. Following the sub-sampling strategy in \cite{bailey2021gender}, we randomly sub-sampled the other three classes to match the smallest class (depressed males with 19 samples), resulting in 19 samples per class in the training set.

The second method is the MixFeat data augmentation technique proposed in \cite{cheong2023s}. Unlike directly augmenting raw samples, MixFeat generates new virtual samples by linearly interpolating feature vectors of minority groups. In our experiments, the augmentation is directly applied on ALF extracted by the backbone model. Let $C_i$ and $C_j$ denote the ALF for the training samples $i$ and $j$, the augmented feature $C_k$ is generated according to the following formula:

\begin{equation}
    C_k = \lambda \cdot C_i + (1 - \lambda) \cdot C_j   
\end{equation}

where samples $i$ and $j$ ($i \neq j$) are drawn from the same target class (e.g., both are depressed males), and the mixing coefficient $\lambda \sim$ Beta(0, 1).

\section{Results and Discussion}
We evaluated our proposed counterfactual debiasing framework against two established debiasing strategies: sub-sampling and data augmentation. The experiments were conducted on the DAIC-WOZ dataset using two distinct architectures: the STA-based model and the NetVLAD-based model. Detailed performance metrics are reported in Table~\ref{tab3}. Note that "None" refers to testing the backbone model directly without applying any debiasing methods. It serves as a reference to quantify the severity of inherent model's gender bias.

In terms of \textbf{overall performance}, our method clearly outperforms existing strategies. As shown in Table~\ref{tab3}, the models without any debiasing methods ("None") reveal a notable performance gap, especially in the STA-based model where Male-F1 (0.582) is lower than Female-F1 (0.644). Although sub-sampling balances the data, it significantly reduces training information, causing the overall F1-score to fall from 0.626 to 0.593. In contrast, our framework enhances the model’s ability to detect true pathological features without losing data. For the STA-based model, we achieve the highest accuracy (0.702) and F1-score (0.644). Importantly, this improvement comes from recovering the minority group’s performance: Male-F1 rises from 0.582 to 0.654, showing the model no longer depends on gender shortcuts. A similar pattern appears in the NetVLAD-based model, where our method sets new records across all metrics, pushing Male-F1 to 0.808 and Female-F1 to 0.798.

Regarding \textbf{fairness}, the metrics confirm that our approach effectively severs the spurious causal link between gender and depression. The models without any debiasing methods ("None") show strong bias, with the STA-based model having a DI of 2.635, which means it tends to predict depression more often for female samples. While Sub-sampling brings DI closer to 1 (0.958), it does so at the cost of overall accuracy (EA improves slightly but F1 drops). Data Augmentation, on the other hand, fails to consistently improve fairness, even worsening EA to 0.056 on the STA-based model.  Our counterfactual framework achieves the best balance: it significantly reduces the accuracy gap (EA drops to 0.013 and 0.007 for STA-based and NetVLAD-based models, respectively) and adjusts DI to a more equitable range (0.719 and 0.745). This shows that our method effectively reduces the model’s dependence on gender information while maintaining its ability to accurately detect actual depression cues.

\begin{table*}[!tbp]
    \centering
    \caption{Performance Results of Different Debiasing Methods on DAIC-WOZ. “None” refers to testing the backbone model directly without applying any debiasing methods.}
    \setlength{\tabcolsep}{4pt}
    \begin{tabular}{c l c c c c c c c}
        \toprule
        \multicolumn{1}{c}{Model} & Debiasing Method & F1-score & Accuracy & Recall & Male-F1 & Female-F1 & EA & DI \\
        \midrule
        \multirow{4}{*}{STA-based Model} & None & 0.626 & 0.681 & 0.629 & 0.582 & 0.644 & 0.029 & 2.635 \\
        & Sub-sampling \cite{bailey2021gender} & 0.593 & 0.660 & 0.593 & 0.589 & 0.597 & 0.015 & \textbf{0.958} \\
        & Data Augmentation \cite{cheong2023s} & 0.639 & 0.681 & \textbf{0.649} & 0.589 & \textbf{0.681} & 0.056 & 1.369 \\
        & Counterfactual Inference (Ours) & \textbf{0.644} & \textbf{0.702} & 0.644 & \textbf{0.654} & 0.631 & \textbf{0.013} & 0.719 \\
        \midrule
        \multirow{4}{*}{NetVLAD-based Model} & None & 0.776 & 0.809 & 0.781 & 0.775 & 0.772 & 0.034 & 1.917 \\
        & Sub-sampling \cite{bailey2021gender} & 0.726 & 0.766 & 0.731 & 0.753 & 0.698 & 0.033 & \textbf{0.839} \\
        & Data Augmentation \cite{cheong2023s} & 0.783 & 0.809 & 0.802 & 0.795 & 0.772 & 0.034 & 1.369 \\
        & Counterfactual Inference (Ours) & \textbf{0.804} & \textbf{0.830} & \textbf{0.817} & \textbf{0.808} & \textbf{0.798} & \textbf{0.007} & 0.745 \\
        \bottomrule
    \end{tabular}
    \label{tab3}
\end{table*}

\section{Conclusion}
In this paper, we identified and addressed the critical issue of gender bias in audio-based depression detection. By modeling the detection process through a causal graph, we revealed that conventional models tend to rely on gender information as a confounder, leading to unfair predictions. To mitigate this, we proposed a counterfactual debiasing framework that quantifies the direct causal effect of gender and removes it during the inference stage via counterfactual intervention. Extensive experiments on the DAIC-WOZ dataset validated that our approach not only significantly improves fairness metrics (EA and DI) but also enhances overall detection accuracy and recall for both genders. For future work, we plan to extend this causal framework beyond the audio modality. Depression diagnosis in clinical settings is inherently multimodal, involving visual and linguistic cues. Since gender bias may vary across these modalities, we aim to develop a unified multimodal counterfactual framework to disentangle and reduce bias from diverse data sources, enabling more robust and fair automated mental health assessments.

\section*{Acknowledgment}
This research was supported by National Natural Science Foundation of China (Grant No. 62502396) and XJTLU Research Development Fund (RDF-24-02-008).

\bibliographystyle{IEEEtran}
\bibliography{ref}

\begin{thebibliography}{10}
\providecommand{\url}[1]{#1}
\csname url@samestyle\endcsname
\providecommand{\newblock}{\relax}
\providecommand{\bibinfo}[2]{#2}
\providecommand{\BIBentrySTDinterwordspacing}{\spaceskip=0pt\relax}
\providecommand{\BIBentryALTinterwordstretchfactor}{4}
\providecommand{\BIBentryALTinterwordspacing}{\spaceskip=\fontdimen2\font plus
\BIBentryALTinterwordstretchfactor\fontdimen3\font minus \fontdimen4\font\relax}
\providecommand{\BIBforeignlanguage}[2]{{%
\expandafter\ifx\csname l@#1\endcsname\relax
\typeout{** WARNING: IEEEtran.bst: No hyphenation pattern has been}%
\typeout{** loaded for the language `#1'. Using the pattern for}%
\typeout{** the default language instead.}%
\else
\language=\csname l@#1\endcsname
\fi
#2}}
\providecommand{\BIBdecl}{\relax}
\BIBdecl

\bibitem{goldman1999awareness}
L.~S. Goldman, N.~H. Nielsen, H.~C. Champion, and A.~M.~A. Council~on Scientific~Affairs, ``Awareness, diagnosis, and treatment of depression,'' \emph{Journal of general internal medicine}, vol.~14, no.~9, pp. 569--580, 1999.

\bibitem{who2025report}
{World Health Organization}, \emph{World mental health today: latest data}.\hskip 1em plus 0.5em minus 0.4em\relax Geneva: World Health Organization, 2025.

\bibitem{kroenke2009phq}
K.~Kroenke, T.~W. Strine, R.~L. Spitzer, J.~B. Williams, J.~T. Berry, and A.~H. Mokdad, ``The phq-8 as a measure of current depression in the general population,'' \emph{Journal of affective disorders}, vol. 114, no. 1-3, pp. 163--173, 2009.

\bibitem{beck1961inventory}
A.~T. Beck, C.~H. Ward, M.~Mendelson, J.~Mock, and J.~Erbaugh, ``An inventory for measuring depression,'' \emph{Archives of general psychiatry}, vol.~4, no.~6, pp. 561--571, 1961.

\bibitem{endicott1978diagnostic}
J.~Endicott and R.~L. Spitzer, ``A diagnostic interview: the schedule for affective disorders and schizophrenia,'' \emph{Archives of general psychiatry}, vol.~35, no.~7, pp. 837--844, 1978.

\bibitem{baevski2020wav2vec}
A.~Baevski, Y.~Zhou, A.~Mohamed, and M.~Auli, ``wav2vec 2.0: A framework for self-supervised learning of speech representations,'' \emph{Advances in neural information processing systems}, vol.~33, pp. 12\,449--12\,460, 2020.

\bibitem{kuehner2017depression}
C.~Kuehner, ``Why is depression more common among women than among men?'' \emph{The lancet psychiatry}, vol.~4, no.~2, pp. 146--158, 2017.

\bibitem{floyd1997problems}
B.~J. Floyd, ``Problems in accurate medical diagnosis of depression in female patients,'' \emph{Social science \& medicine}, vol.~44, no.~3, pp. 403--412, 1997.

\bibitem{shi2021hypothesis}
P.~Shi, A.~Yang, Q.~Zhao, Z.~Chen, X.~Ren, and Q.~Dai, ``A hypothesis of gender differences in self-reporting symptom of depression: implications to solve under-diagnosis and under-treatment of depression in males,'' \emph{Frontiers in psychiatry}, vol.~12, p. 589687, 2021.

\bibitem{bailey2021gender}
A.~Bailey and M.~D. Plumbley, ``Gender bias in depression detection using audio features,'' in \emph{2021 29th European Signal Processing Conference (EUSIPCO)}.\hskip 1em plus 0.5em minus 0.4em\relax IEEE, 2021, pp. 596--600.

\bibitem{cheong2023s}
J.~Cheong, M.~Spitale, and H.~Gunes, ``“it’s not fair!”--fairness for a small dataset of multi-modal dyadic mental well-being coaching,'' in \emph{2023 11th International Conference on Affective Computing and Intelligent Interaction (ACII)}.\hskip 1em plus 0.5em minus 0.4em\relax IEEE, 2023, pp. 1--8.

\bibitem{pearl2010introduction}
J.~\vspace{0mm}Pearl, ``An introduction to causal inference,'' \emph{The international journal of biostatistics}, vol.~6, no.~2, p.~7, 2010.

\bibitem{pearl2011logic}
J.~Pearl, ``\vspace{0mm}the logic of counterfactuals in causal inference,'' 2011.

\bibitem{niu2021counterfactual}
Y.~Niu, K.~Tang, H.~Zhang, Z.~Lu, X.-S. Hua, and J.-R. Wen, ``Counterfactual vqa: A cause-effect look at language bias,'' in \emph{Proceedings of the IEEE/CVF conference on computer vision and pattern recognition}, 2021, pp. 12\,700--12\,710.

\bibitem{li2025automated}
Y.~Li, S.~Kumbale, Y.~Chen, T.~Surana, E.~S. Chng, and C.~Guan, ``Automated depression detection from text and audio: A systematic review,'' \emph{IEEE Journal of Biomedical and Health Informatics}, 2025.

\bibitem{huang2019natural}
Z.~Huang, J.~Epps, D.~Joachim, and V.~Sethu, ``Natural language processing methods for acoustic and landmark event-based features in speech-based depression detection,'' \emph{IEEE Journal of selected topics in Signal Processing}, vol.~14, no.~2, pp. 435--448, 2019.

\bibitem{jayawardena2020ordinal}
S.~Jayawardena, J.~Epps, and E.~Ambikairajah, ``Ordinal logistic regression with partial proportional odds for depression prediction,'' \emph{IEEE Transactions on Affective Computing}, vol.~14, no.~1, pp. 563--577, 2020.

\bibitem{xu2025deep}
C.~Xu, Y.~Chen, Y.~Tao, W.~Xie, X.~Liu, Y.~Lin, C.~Liang, F.~Du, Z.~Zhi, and C.~Shi, ``Deep learning-based detection of depression by fusing auditory, visual and textual clues,'' \emph{Journal of Affective Disorders}, p. 119860, 2025.

\bibitem{chen2025speech}
Y.~Chen, C.~Xu, C.~Liang, Y.~Tao, and C.~Shi, ``Speech-based clinical depression screening: an empirical study,'' in \emph{IEEE International Conference on Acoustics, Speech, and Signal Processing}, 2025.

\bibitem{niu2020multimodal}
M.~Niu, J.~Tao, B.~Liu, J.~Huang, and Z.~Lian, ``Multimodal spatiotemporal representation for automatic depression level detection,'' \emph{IEEE transactions on affective computing}, vol.~14, no.~1, pp. 294--307, 2020.

\bibitem{shen2022automatic}
Y.~Shen, H.~Yang, and L.~Lin, ``Automatic depression detection: An emotional audio-textual corpus and a gru/bilstm-based model,'' in \emph{ICASSP 2022-2022 IEEE International Conference on Acoustics, Speech and Signal Processing (ICASSP)}.\hskip 1em plus 0.5em minus 0.4em\relax IEEE, 2022, pp. 6247--6251.

\bibitem{cheong2021hitchhiker}
J.~Cheong, S.~Kalkan, and H.~Gunes, ``The hitchhiker’s guide to bias and fairness in facial affective signal processing: Overview and techniques,'' \emph{IEEE Signal Processing Magazine}, vol.~38, no.~6, pp. 39--49, 2021.

\bibitem{zhang2024novel}
Z.~Zhang, Q.~Meng, L.~Jin, H.~Wang, and H.~Hou, ``A novel eeg-based graph convolution network for depression detection: incorporating secondary subject partitioning and attention mechanism,'' \emph{Expert Systems with Applications}, vol. 239, p. 122356, 2024.

\bibitem{dang2024fairness}
V.~N. Dang, A.~Cascarano, R.~H. Mulder, C.~Cecil, M.~A. Zuluaga, J.~Hern{\'a}ndez-Gonz{\'a}lez, and K.~Lekadir, ``Fairness and bias correction in machine learning for depression prediction across four study populations,'' \emph{Scientific Reports}, vol.~14, no.~1, p. 7848, 2024.

\bibitem{liu2025large}
X.~Liu, P.~Xu, J.~Wu, J.~Yuan, Y.~Yang, Y.~Zhou, F.~Liu, T.~Guan, H.~Wang, T.~Yu \emph{et~al.}, ``Large language models and causal inference in collaboration: A comprehensive survey,'' \emph{Findings of the Association for Computational Linguistics: NAACL 2025}, pp. 7668--7684, 2025.

\bibitem{luo2024survey}
H.~Luo, F.~Zhuang, R.~Xie, H.~Zhu, D.~Wang, Z.~An, and Y.~Xu, ``A survey on causal inference for recommendation,'' \emph{The Innovation}, vol.~5, no.~2, 2024.

\bibitem{gratch2014distress}
J.~Gratch, R.~Artstein, G.~M. Lucas, G.~Stratou, S.~Scherer, A.~Nazarian, R.~Wood, J.~Boberg, D.~DeVault, S.~Marsella \emph{et~al.}, ``The distress analysis interview corpus of human and computer interviews.'' in \emph{Lrec}, vol.~14.\hskip 1em plus 0.5em minus 0.4em\relax Reykjavik, 2014, pp. 3123--3128.

\end{thebibliography}

\end{document}